\documentclass[conference]{IEEEtran}
\IEEEoverridecommandlockouts
\usepackage{cite}
\usepackage{amsmath,amssymb,amsfonts}
\usepackage{algorithmic}
\usepackage{graphicx}
\usepackage{textcomp}
\usepackage{xcolor}
\definecolor{lightblue}{RGB}{30,144,255}
\usepackage[hidelinks]{hyperref}
\hypersetup{
    colorlinks=true,
    linkcolor=lightblue,
    citecolor=lightblue,
    urlcolor=lightblue
}
\usepackage{subcaption}
\usepackage{stfloats}
\usepackage{fancyhdr}
\usepackage{booktabs}
\def\BibTeX{{\rm B\kern-.05em{\sc i\kern-.025em b}\kern-.08em
    T\kern-.1667em\lower.7ex\hbox{E}\kern-.125emX}}
\begin{document}

\title{
% Rig-Aware Gaussian Splatting for Wide-Angle, Multi-View Under-Vehicle Inspection
% A Rig-Aware SfM Pipeline with LightGlue for Gaussian Splatting of Vehicle Undercarriages
% Photorealistic 3D Reconstruction of Vehicle Undercarriages using Multi-View Gaussian Splatting with Strong Geometric Priors

% Overcoming Low-Parallax and Wide-Angle Challenges in SfM with a Rig-Aware, Learning-Based Pipeline for 3D Vehicle Inspection

% Interactive 3D Under-Vehicle Inspection via Rig-Aware Gaussian Splatting

% Rig-Aware Gaussian Splatting for 3D Reconstruction of Vehicle Undercarriages

% Rig-Aware Structure-from-Motion Pipeline for Gaussian Splatting of Vehicle Undercarriages

Rig-Aware 3D Reconstruction of Vehicle Undercarriages using Gaussian Splatting

% {\footnotesize \textsuperscript{*}Note: Sub-titles are not captured in Xplore and
% should not be used}
% \thanks{Identify applicable funding agency here. If none, delete this.}
% \vspace{-0.25 cm}
}

\author{%
  \IEEEauthorblockN{%
    Nitin Kulkarni\IEEEauthorrefmark{1}\IEEEauthorrefmark{2}, 
    Akhil Devarashetti\IEEEauthorrefmark{1}, 
    Charlie Cluss\IEEEauthorrefmark{1}, 
    Livio Forte\IEEEauthorrefmark{1},\\
    Dan Buckmaster\IEEEauthorrefmark{1}, 
    Philip Schneider\IEEEauthorrefmark{1}, 
    Chunming Qiao\thanks{C. Qiao is supported in part by the National Science Foundation under Grant Nos. CNS-2413876 and CNS-2120369. The authors thank ACV Auctions for providing computing resources, vehicle data, and the camera rig.}\IEEEauthorrefmark{2}, 
    Alina Vereshchaka\IEEEauthorrefmark{2}%
  }
  \IEEEauthorblockA{\IEEEauthorrefmark{2}University at Buffalo, Buffalo, NY, USA}
  \IEEEauthorblockA{\IEEEauthorrefmark{1}ACV Auctions, Buffalo, NY, USA}
  \IEEEauthorblockA{%
  \{nitinvis, qiao, avereshc\}@buffalo.edu, \{adevarashetti, ccluss, lforte, dbuckmaster, pschneider\}@acvauctions.com}
}

\maketitle

\thispagestyle{fancy}
\fancyhf{} % Clear header/footer
\renewcommand{\headrulewidth}{0pt} % Remove header rule
\fancyfoot[c]{%
  \parbox{\textwidth}{%
  \centering \scriptsize
  \copyright~2025 IEEE. Personal use of this material is permitted. Permission from IEEE must be obtained for all other uses, in any current or future media, including reprinting/republishing this material for advertising or promotional purposes, creating new collective works, for resale or redistribution to servers or lists, or reuse of any copyrighted component of this work in other works.
  }
}

\begin{abstract}
Inspecting the undercarriage of used vehicles is a labor-intensive task that requires inspectors to crouch or crawl underneath each vehicle to thoroughly examine it. Additionally, online buyers rarely see undercarriage photos. We present an end-to-end pipeline that utilizes a three-camera rig to capture videos of the undercarriage as the vehicle drives over it, and produces an interactive 3D model of the undercarriage. The 3D model enables inspectors and customers to rotate, zoom, and slice through the undercarriage, allowing them to detect rust, leaks, or impact damage in seconds, thereby improving both workplace safety and buyer confidence. Our primary contribution is a rig-aware Structure-from-Motion (SfM) pipeline specifically designed to overcome the challenges of wide-angle lens distortion and low-parallax scenes. Our method overcomes the challenges of wide-angle lens distortion and low-parallax scenes by integrating precise camera calibration, synchronized video streams, and strong geometric priors from the camera rig. We use a constrained matching strategy with learned components, the DISK feature extractor, and the attention-based LightGlue matcher to generate high-quality sparse point clouds that are often unattainable with standard SfM pipelines. These point clouds seed the Gaussian splatting process to generate photorealistic undercarriage models that render in real-time. Our experiments and ablation studies demonstrate that our design choices are essential to achieve state-of-the-art quality.

% Our primary contribution is a rig-aware Structure-from-Motion (SfM) pipeline specifically designed to overcome the challenges of wide-angle lens distortion and low-parallax scenes. By incorporating precise camera calibration, spatial-temporal video synchronization, strong geometric priors from the camera rig, and a constrained feature matching strategy that utilizes state-of-the-art learned components, including the DISK feature extractor and the attention-based LightGlue matcher, our method generates high-quality sparse point clouds that other SfM approaches often fail to produce. These point clouds seed the Gaussian Splatting process to generate photorealistic undercarriage models that render in real-time. Our comprehensive experiments and ablation studies demonstrate that our design choices are critical to achieve state-of-the-art quality.

\end{abstract}

\begin{IEEEkeywords} Gaussian splatting, structure-from-motion, LightGlue, NeRFs, 3D reconstruction, photogrammetry, vehicle inspection
\end{IEEEkeywords}

\section{Introduction}
Used vehicles represent a multi-billion-dollar global market~\cite{chauhan2025used}, with digital auctions capturing an increasingly large share. Yet, assessing a vehicle's underbody condition, one of the key factors in determining its value, remains challenging. Signs of rust, fluid leaks, or frame damage are often obscured from standard imaging and inspection routines. Vehicle inspectors may miss issues due to cramped working conditions, and buyers, especially in rural or remote areas, often lack access to trusted inspection services. This challenge is particularly pronounced for individuals in rural or remote areas, who often lack access to specialized inspection services. Without clear visibility into the vehicle's underside, buyers make decisions under uncertainty, risking costly repairs and reducing trust in online transactions. 

To bridge this information gap, we propose a solution that delivers high-resolution, photorealistic, and interactive 3D models of vehicle undercarriages. Such 3D models enable buyers to remotely inspect a vehicle with fine-grained detail. For sellers and inspectors, this pipeline provides increased transparency and accountability, ultimately improving trust and setting a new standard in the resale marketplace.

Our system captures high-resolution videos from three laterally spaced wide-angle cameras as a vehicle drives over a custom rig. Wide-angle lenses (160° FOV) are required to capture the undercarriage as the distance between the cameras and the undercarriage is short ($12 - 30 \,\rm{cm}$). However, such wide-angle lenses introduce significant distortion, which makes feature matching difficult. A single camera lens only provides parallax along the length of the vehicle as the vehicle moves in a straight line over the cameras. To generate width-wise parallax for depth estimation, we use three laterally spaced cameras with wide-angle lenses. Traditional Structure-from-Motion (SfM) pipelines often struggle under these conditions, and Neural Radiance Fields (NeRFs) \cite{mildenhall2021nerf} are too slow for production-level vehicle inspection pipelines.

We introduce a novel, modular pipeline that integrates traditional multi-view geometry with 3D Gaussian splatting (3D-GS) \cite{kerbl20233d} to efficiently reconstruct detailed and interactive undercarriage models. First, we perform precise camera calibration using ChArUco boards and select a diverse and sharp set of images to model lens distortion and minimize projection error (Sec.~\ref{sub:camera_calibration}). Next, we synchronize multi-camera streams at sub-frame precision by estimating global vertical shifts using phase correlation, followed by refining the timing offsets through adaptive cross-correlation (Sec.~\ref{sub:video_synchronization}). This two-stage alignment achieves millisecond-level synchronization, which is needed for our downstream processing.

% We then apply our rig-aware SfM approach that uses state-of-the-art learned components. After selecting the sharpest image triplets, we use our calibrated intrinsics to undistort the images. From these undistorted images, we extract dense local features using DISK \cite{tyszkiewicz2020disk}, a learned keypoint detector designed to work well in challenging conditions. We match features using the attention-based matcher LightGlue \cite{lindenberger2023lightglue} on a constrained set of spatiotemporally proximate image pairs. The resulting raw matches are geometrically verified within COLMAP \cite{schoenberger2016sfm}. Finally, we use GLOMAP \cite{pan2024global} to run global bundle adjustment, using strong priors from the known camera rig geometry, to refine the camera poses and generate a clean and accurate point cloud of the undercarriage (Sec.~\ref{sub:structure_from_motion}). The point cloud and optimized camera poses serve as the seed for Gaussian Splatting to generate the interactive 3D model (Sec.~\ref{sub:gaussian_splatting}).

% Removed the word strong from "strong priors". Reworded, "Finally, we use GLOMAP" to "We then use GLOMAP".
% \textcolor{red}{
We then apply our rig-aware SfM approach that uses state-of-the-art learned components. After selecting the sharpest image triplets, we use our calibrated intrinsics to undistort the images. From these undistorted images, we extract dense local features using DISK \cite{tyszkiewicz2020disk}, a learned keypoint detector designed to work well in challenging conditions. We match features using the attention-based matcher LightGlue \cite{lindenberger2023lightglue} on a constrained set of spatiotemporally proximate image pairs. The resulting raw matches are geometrically verified within COLMAP \cite{schoenberger2016sfm}. We then use GLOMAP \cite{pan2024global} to run global bundle adjustment, using priors from the known camera rig geometry, to refine the camera poses and generate a clean and accurate point cloud of the undercarriage (Sec.~\ref{sub:structure_from_motion}). The point cloud and optimized camera poses serve as the seed for Gaussian splatting to generate the interactive 3D model (Sec.~\ref{sub:gaussian_splatting}).
% }

Our proposed pipeline efficiently generates inspection-grade 3D reconstructions, allowing vehicle inspectors and buyers to inspect a car’s undercarriage interactively in real-time. We demonstrate that precise calibration, synchronization, and geometry-aware SfM can achieve NeRF-level quality on challenging wide-angle, low-parallax sequences, making this a practical and scalable solution for real-world deployment. This technology has the potential to standardize vehicle inspections across online marketplaces, enhance trust, and support applications such as insurance claims and fleet maintenance.

\section{Background}
% Creating accurate, interactive 3D models of real-world scenes from a collection of 2D images is a long-standing goal in computer vision. This process, known as novel view synthesis, lets us generate new views of a scene as if the camera had been placed in a different position. It is fundamental to applications such as virtual reality, digital heritage \cite{snavely2006photo}, and industrial inspection. Our work builds on key strands of 3D reconstruction: classical methods based on multi-view geometry, modern deep-learning-based approaches for feature extraction and matching, and recent advances in radiance fields.

% Accurate novel-view synthesis from 2D images is fundamental to applications such as virtual reality, cultural-heritage digitization, and industrial inspection \cite{snavely2006photo}. Our pipeline integrates three research strands: classical multi-view geometry, learned feature matching, and fast radiance-field–style rendering.

Accurate novel-view synthesis from 2D images is fundamental to applications such as virtual reality, cultural-heritage digitization, and industrial inspection \cite{snavely2006photo}. Our work builds on key strands of 3D reconstruction: classical multi-view geometry, deep learning models for feature extraction and matching, and recent advances in radiance fields.

\subsection{Structure-from-Motion}
Structure-from-Motion (SfM) estimates 3D scene geometry and camera poses from overlapping images by detecting and matching keypoints across views. Starting from an initial image pair, systems like COLMAP incrementally add new views through resectioning and triangulation, followed by bundle adjustment (BA) to jointly refine camera poses and 3D point locations. The result is a sparse 3D point cloud and a globally consistent camera trajectory.

However, standard SfM pipelines face challenges with distortion and low-parallax scenes, and the quality of the reconstruction is highly dependent on the quality and distribution of feature matches. Scenes captured with wide-angle lenses, as required in our undercarriage inspection, introduce severe nonlinear distortions that can confound feature matching if not accurately modeled \cite{mei2007single}. Additionally, SfM struggles with camera configurations with low parallax;  differences between views that are crucial for depth estimation \cite{hartley2003multiple}. This scenario inherent to our drive-over camera rig can lead to geometric degeneracies and drift in the estimated camera poses, ultimately degrading the quality of the final 3D point cloud.

\subsection{Learned Local Features and Matchers}
Traditional SfM pipelines have relied on handcrafted features like SIFT \cite{lowe2004distinctive}, but recent progress in deep learning has led to more powerful alternatives. Learned local features, such as DISK, are trained on large datasets of images to detect keypoints and generate descriptors that are more robust to extreme viewpoint and illumination changes than classical methods.

% % Removed the word "recent" from "recent progress in deep learning". Removed the word "more" from "has led to more powerful alternatives".
% \textcolor{red}{
% Traditional SfM pipelines have relied on handcrafted features like SIFT \cite{lowe2004distinctive}, but progress in deep learning has led to powerful alternatives. Learned local features, such as DISK, are trained on large datasets of images to detect keypoints and generate descriptors that are more robust to extreme viewpoint and illumination changes than classical methods.
% }

Similarly, deep learning-based matchers have enhanced the ability to match features across images. Instead of relying on nearest-neighbor matching, modern approaches, such as LoFTR \cite{sun2021loftr}, SuperGlue \cite{sarlin2020superglue}, and LightGlue \cite{lindenberger2023lightglue}, utilize attention-based graph neural networks to consider the global context of all features in both images simultaneously. This helps them find accurate matches even in low-texture or repetitive scenes that are difficult for traditional methods. Our work integrates these learned feature extractors and matchers into a geometrically constrained SfM pipeline to overcome the challenges of our setup.

% % Removed the word "simple" from "Instead of relying on simple nearest-neighbor matching...". Replaced "This helps them find accurate matches even in..." with "This results in accurate matches even in...". 
% \textcolor{red}{
% Similarly, deep learning based matchers have enhanced our ability to match features across images. Instead of relying on nearest-neighbor matching, modern approaches, such as SuperGlue \cite{sarlin2020superglue} and LightGlue \cite{lindenberger2023lightglue}, utilize attention-based graph neural networks to consider the global context of all features in both images simultaneously. This results in accurate matches even in low-texture or repetitive scenes that are difficult for traditional methods. Our work integrates these learned feature extractors and matchers into a geometrically constrained SfM pipeline to overcome the challenges of our setup.
% }

\subsection{Radiance Fields and Novel View Synthesis}
While SfM captures sparse geometry, it does not provide photorealistic renderings. Earlier pipelines filled this gap with Multi-View Stereo (MVS) \cite{goesele2006multi}, surface reconstruction, and texture mapping. More recently, Neural Radiance Fields (NeRFs) \cite{mildenhall2021nerf} have shown exceptional capability in synthesizing novel views by learning a volumetric scene representation with a neural network.

% NeRFs can produce realistic 3D scenes that capture fine details, such as reflections and translucency. However, they are computationally expensive, often requiring many hours to train and rendering at only a few frames per minute. This performance bottleneck renders NeRF impractical for applications requiring real-time interaction and rapid processing, such as our vehicle inspection workflow.
NeRFs can produce realistic 3D scenes that capture fine details, such as reflections and translucency. However, they are computationally expensive, often requiring many hours to train and rendering at only a few frames per minute. Subsequent NeRF variants reduced per-scene training time; for example, Instant-NGP uses a multi-resolution hash encoding and optimized kernels for faster training and interactive rendering \cite{muller2022instant}. 3D Gaussian splatting (3D-GS), by contrast, is an explicit SfM-seeded representation that models scenes as anisotropic Gaussians and uses visibility-aware splatting to enable fast optimization and real-time rendering, a trade-off well suited to our drive-over undercarriage workflow \cite{kerbl20233d, fei20243d}.

% While NeRF produces exceptionally high-quality novel views, training and rendering are slow (training often takes hours for a single scene). Subsequent work in the NeRF family dramatically reduces training time, e.g., Instant-NGP uses a multiresolution hash encoding and highly optimized kernels to obtain near-instant training and interactive rendering in many settings \cite{muller2022instant}. However, these faster implicit methods remain fundamentally different from explicit, rasterization-based representations: 3D Gaussian Splatting (3D-GS) represents the scene as optimized anisotropic Gaussians initialized from SfM points, and leverages a visibility-aware splatting/rasterization pipeline to achieve fast optimization and real-time rendering with strong visual fidelity. This explicit representation and its SfM initialization are particularly attractive for our drive-over undercarriage workflow, where fast training, accurate alignment to sparse SfM geometry, and interactive inspection are essential \cite{kerbl20233d, fei20243d}.

\subsection{Gaussian Splatting}
3D Gaussian splatting (3D-GS) \cite{kerbl20233d} offers a real-time alternative to NeRFs, achieving comparable or better visual fidelity with faster training and rendering. Instead of an implicit neural representation, 3D-GS models the scene explicitly as a large collection of 3D Gaussians.

Each Gaussian is defined by a set of optimizable attributes: its position (mean $\mu$), shape and orientation (a $3 \times 3$ covariance matrix $\Sigma$), color (represented by spherical harmonics), and opacity ($\alpha$). These Gaussians are initialized from the sparse point cloud generated from SfM. During training, their attributes are optimized to reconstruct the training images. 

Unlike NeRF, 3D-GS does not need to trace rays through the scene. Instead, it projects each Gaussian directly onto the images, like soft 2D dots (``splats") and combines them using alpha blending. This is done efficiently using a fast rasterizer. This results in fast training time and the ability to render in real-time. 

As 3D-GS relies on an initial SfM point cloud, the quality of the initial reconstruction is critical. This directly motivates our rig-aware SfM pipeline, designed to handle low-parallax, wide-angle scenes like vehicle undercarriages.

\section{Methodology}

\begin{figure*}[!t]
  \centering
  \includegraphics[width=\textwidth]{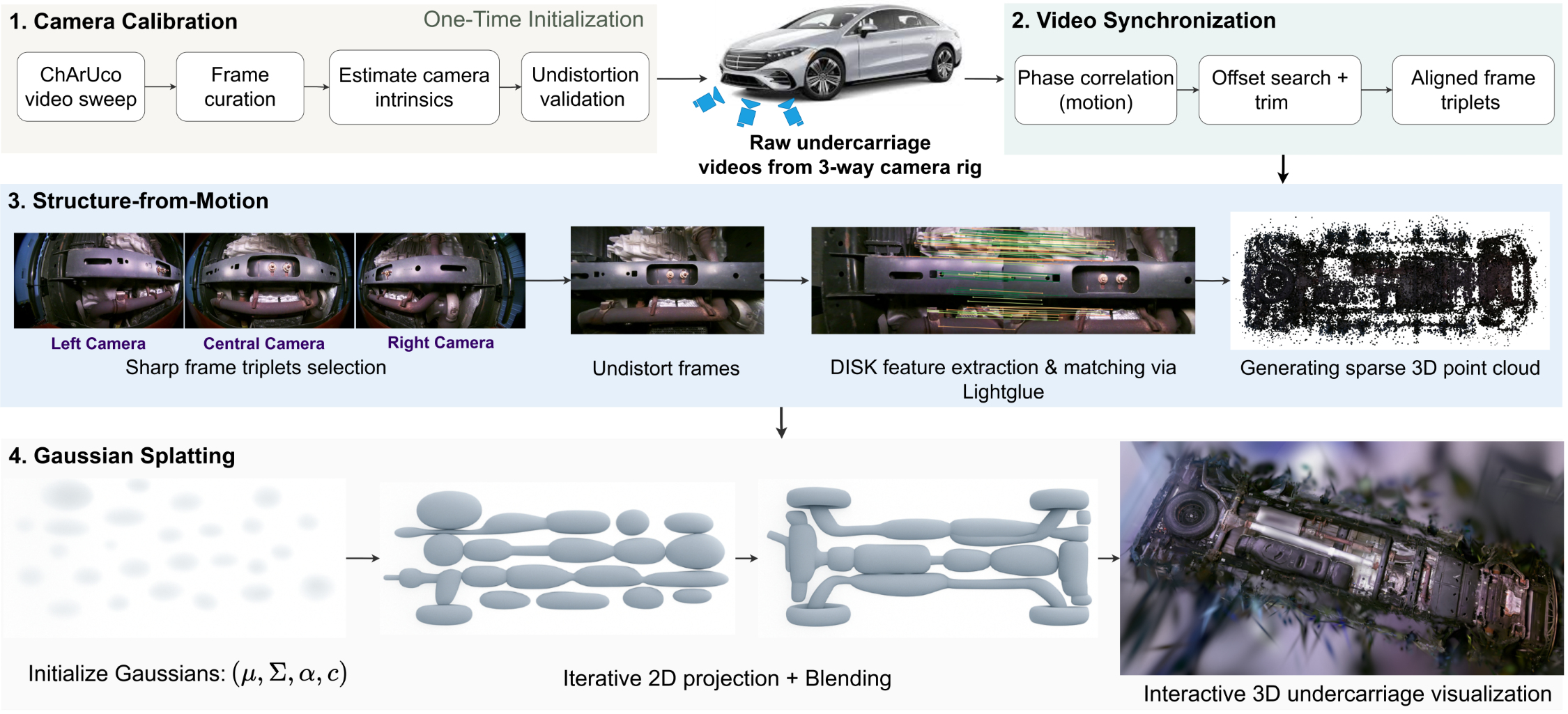}
  \caption{3D reconstruction pipeline overview. (1) We perform a one-time camera calibration using a ChArUco board to model and correct for severe wide-angle lens distortion. (2) For each vehicle, we synchronize the raw videos from the three cameras on our rig to ensure spatial-temporal alignment. (3) From the synchronized videos, we uniformly sample the sharpest frames triplets, undistort the frames, extract DISK features, apply a constrained feature matching strategy to find matches across different frames via LightGlue, and generate the 3D sparse point cloud of the undercarriage via bundle adjustment. (4) Finally, the point cloud is used to initialize the 3D Gaussians, which are optimized to produce the interactive 3D undercarriage model.}
  \label{fig:framework}
  % \vspace{-0.25 cm}
\end{figure*}

Our pipeline, illustrated in Fig. \ref{fig:framework}, converts raw videos from multiple camera views into a high-quality, interactive 3D model of a vehicle's undercarriage. There are four major steps in our pipeline. The first step (Sec.~\ref{sub:camera_calibration}) is precise camera calibration to correct severe lens distortion. This calibration only needs to be performed once; the calibrated camera parameters can be used to reconstruct 3D models of any vehicle undercarriage. The second step (Sec.~\ref{sub:video_synchronization}) is to synchronize the videos from the three cameras to ensure spatial-temporal correspondence across the three cameras. The third step (Sec. ~\ref{sub:structure_from_motion}) is rig-aware Structure-from-Motion (SfM) to estimate the 3D geometry and camera poses. The last step (Sec.~\ref{sub:gaussian_splatting}) is 3D Gaussian splatting to generate a photorealistic, interactive 3D model of the vehicle undercarriage. Each step is designed to address the challenges posed by wide-angle lenses or low-parallax views.

\subsection{Camera Calibration}
\label{sub:camera_calibration}
Wide-angle lenses, while necessary for capturing the undercarriage from a short distance, introduce severe radial and tangential distortion; without accurate camera intrinsic parameters, these non-linear distortions propagate into the SfM stage and warp the sparse point cloud. While SfM frameworks such as COLMAP can estimate camera intrinsic parameters, using a dedicated calibration pipeline produces noticeably lower projection error and reduces bundle-adjustment drift.

\subsubsection{ChArUco Board}
\begin{figure}[t]
    \centering    
    \includegraphics[width=0.78\linewidth]{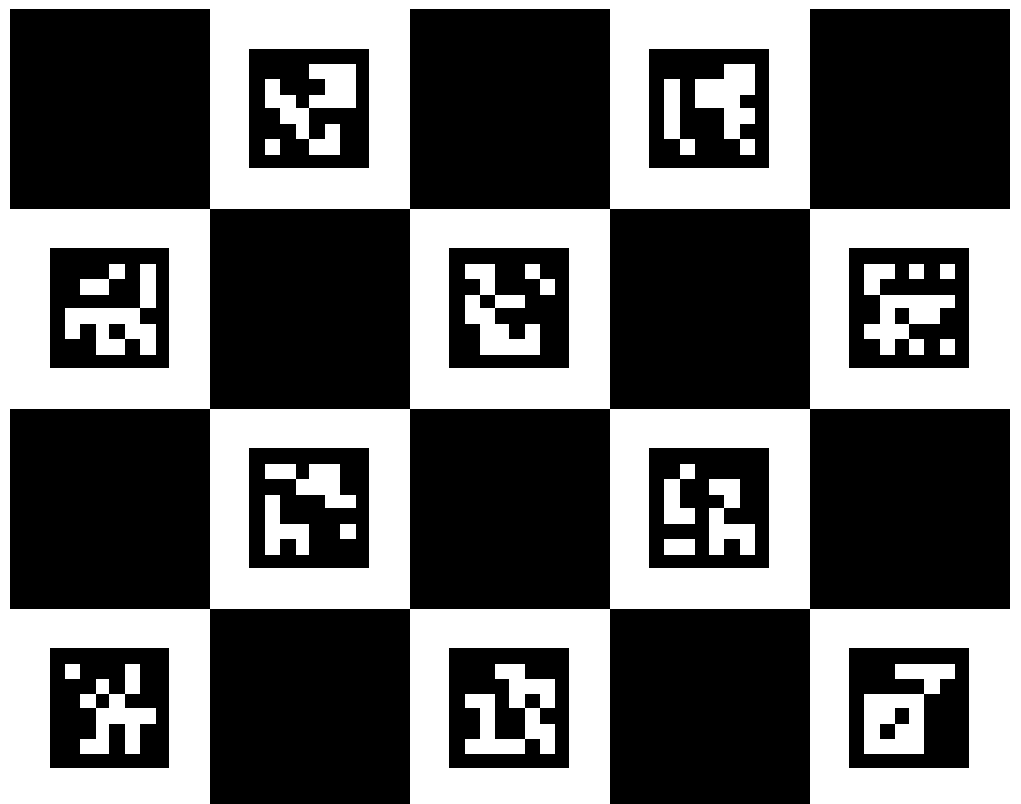}
    \caption{Sample of ChArUco board \cite{opencv_charuco_tutorial}.}   \label{fig:charucoboard}
    % \vspace{-0.4 cm}
\end{figure}

We use a ChArUco board (Fig.~\ref{fig:charucoboard}), a chessboard pattern in which every second square is replaced by an ArUco marker. ChArUco boards provide precise sub-pixel corner detection of chessboard corners and robust ID-based detection of ArUco tags, resulting in lower projection error compared to marker-only or plain chessboard patterns \cite{an2018charuco}. Our board contains $53\times37$ inner squares of side length $22\, \rm{mm}$, with each ArUco marker being $16\, \rm{mm}$ wide. We capture the calibration videos by sweeping the board through the three wide-angle cameras, varying roll, pitch, yaw, and distance ($\approx$~$30 - 120\, \rm{cm}$), ensuring each camera observes large baseline changes.

\subsubsection{Frame Curation}
We sample a diverse and sharp set of calibration images by sliding through a window of ten consecutive images from each video. Within each sliding window, the images are converted to grayscale and scored by the variance of the Laplacian (Eq. \ref{eq:laplacian_variance}). The Laplacian highlights regions of rapid intensity change, making its variance an effective metric for edge acuity and overall image sharpness. 
% Equation from ChatGPT:
\begin{equation}
\label{eq:laplacian_variance}
\mathrm{Var}\bigl(\nabla^2 I\bigr) = \frac{1}{MN} \sum_{x=1}^{M} \sum_{y=1}^{N}
\left(\nabla^2 I(x,y) - \mu_{\nabla^2}\right)^2
\end{equation}
\begin{equation*}
\label{eq:laplacian_mean}
\text{where, } \mu_{\nabla^2} = \frac{1}{MN} \sum_{x=1}^{M} \sum_{y=1}^{N} \nabla^2 I(x,y)
\end{equation*}

Higher values indicate crisper edges and less blur. From each ten-frame window, we sample the frame with the highest Laplacian variance, ensuring we sample the sharpest frames with minimal motion blur. 

\subsubsection{Camera Models and Optimization}
The projection of a 3D point $P=[X,Y,Z]^\top$ to a 2D pixel coordinate $p=[u,v]^\top$ is modeled in two steps. First, the 3D point is projected to normalized, undistorted image coordinates $p' = [x', y']^\top$, where $x' = X/Z$ and $y' = Y/Z$. Second, a distortion function is applied to $p'$ to get distorted coordinates $P_d = [x_d, y_d]^\top$, which are then mapped to pixel coordinates using the camera intrinsic matrix $K$ (Eq.~\ref{eq:camera_matrix}).

{\scriptsize
\begin{equation}
\label{eq:camera_matrix}
\mathbf{K} = \begin{bmatrix} f_x & 0 & c_x \\ 0 & f_y & c_y \\ 0 & 0 & 1 \end{bmatrix}
\end{equation}
}

We fit the eight-parameter Full OpenCV camera model~(Eq.~\ref{eq:full_opencv_distortion}) using the Levenberg-Marquardt optimizer~\cite{ranganathan2004levenberg} to find the distortion parameters for our lenses. This model combines radial distortion (terms with $k_i$) and tangential distortion (terms with $p_i$), which accounts for decentering and non-parallel lens elements. It extends the standard OpenCV model with three additional radial distortion coefficients ($k_4, k_5,k_6$) to better fit the complex distortion profiles of extreme wide-angle lenses.

{\scriptsize
\begin{equation}
\label{eq:full_opencv_distortion}
\begin{aligned}
x_d &= x' \frac{1 + k_1r^2 + k_2r^4 + k_3r^6}{1 + k_4r^2 + k_5r^4 + k_6r^6} + 2p_1x'y' + p_2(r^2+2x'^2) \\
y_d &= y' \frac{1 + k_1r^2 + k_2r^4 + k_3r^6}{1 + k_4r^2 + k_5r^4 + k_6r^6} + p_1(r^2+2y'^2) + 2p_2x'y'
\end{aligned}
\end{equation}
}

\subsubsection{Error Metric} 
The optimization minimizes the Root-Mean-Square (RMS) reprojection error, which measures the Euclidean distance in pixels between the observed corner locations $p_i$ on the ChArUco board and their corresponding projected locations $\hat{p_i}$, as computed by the model: $
\operatorname{RMS}~=~\sqrt{\frac{1}{N}\sum_{i=1}^{N}\lVert\mathbf{p}_i-\hat{\mathbf{p}}_i\rVert^2}
$

% The model yielding the lowest RMS error is selected. This quantitative metric is supplemented by a qualitative evaluation, where we visually inspect the undistorted images to confirm that straight lines in the scene appear straight (Fig. \ref{fig:charuco_calibration_comparison}). The resulting intrinsic matrix K and the distortion coefficients from the best-fit model are then fixed and passed to our SfM pipeline (Sec.~\ref{sub:structure_from_motion}).

This quantitative metric is supplemented by a qualitative evaluation, where we visually inspect the undistorted images to confirm that straight lines in the scene appear straight (Fig. \ref{fig:charuco_calibration_comparison}). The resulting intrinsic matrix K and the distortion coefficients are then fixed and passed to our SfM pipeline (Sec.~\ref{sub:structure_from_motion}).

\subsection{Video Synchronization}
\label{sub:video_synchronization}

Our camera rig is programmed to trigger video recording simultaneously on all three cameras. However, due to signal and encoding latencies, the resulting videos may be slightly out of sync. We address this issue in two stages: first, we estimate global vertical motion, then we align the motion profiles by minimizing the L1 loss between them.

\subsubsection{Global Motion Estimation}
For every pair of consecutive frames, we compute the number of pixels that moved in the vertical direction using phase correlation. This yields a sequence of vertical displacements, or \textit{shifts}, denoted $S \in \mathbb{Z}^{N-1}$ for a video with $N$ frames. We preprocess each frame by applying a Gaussian blur, contrast-limited adaptive histogram equalization (CLAHE), and a Laplacian filter to highlight its key features. 

For the three videos captured by the left, center, and right cameras $\{l, c, r\}$, we obtain three corresponding shift sequences $\{S_l, S_c, S_r\}$ of lengths $\{N_l - 1, N_c - 1, N_r - 1\}$, respectively. These shifts provide a compressed signal summarizing global vertical motion over time, which we use for temporal alignment.

\subsubsection{Offset Search via L1 Loss Minimization}
We synchronize the videos by identifying the temporal offsets that minimize the average absolute difference between their shift signals. For any two videos $a$ and $b$, we define the L1 alignment loss as:

{\scriptsize
\begin{equation}
\mathcal{L}(a, b) = \frac{1}{\overline{N}} \sum_{i=1}^{\overline{N}} \lvert S_i^a - S_i^b \rvert, 
\quad 
\text{where } \overline{N} = \min(N^a - 1, N^b - 1)
\label{eq:sync_loss}
\end{equation}
}

\begin{figure}
     \centering
     \begin{subfigure}{\linewidth}
         \centering
         \includegraphics[width=\textwidth]{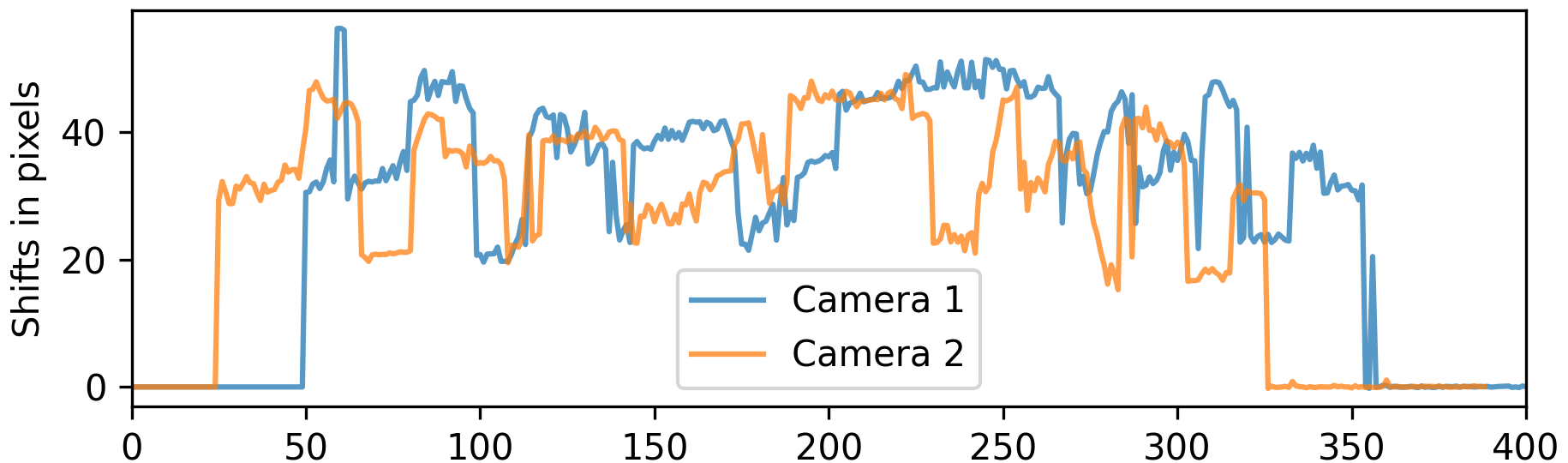}
         \caption{Shifts for 2 videos before syncing.}
         \label{fig:sync1}
     \end{subfigure}
     \hfill
     \begin{subfigure}{\linewidth}
         \centering
         \includegraphics[width=\textwidth]{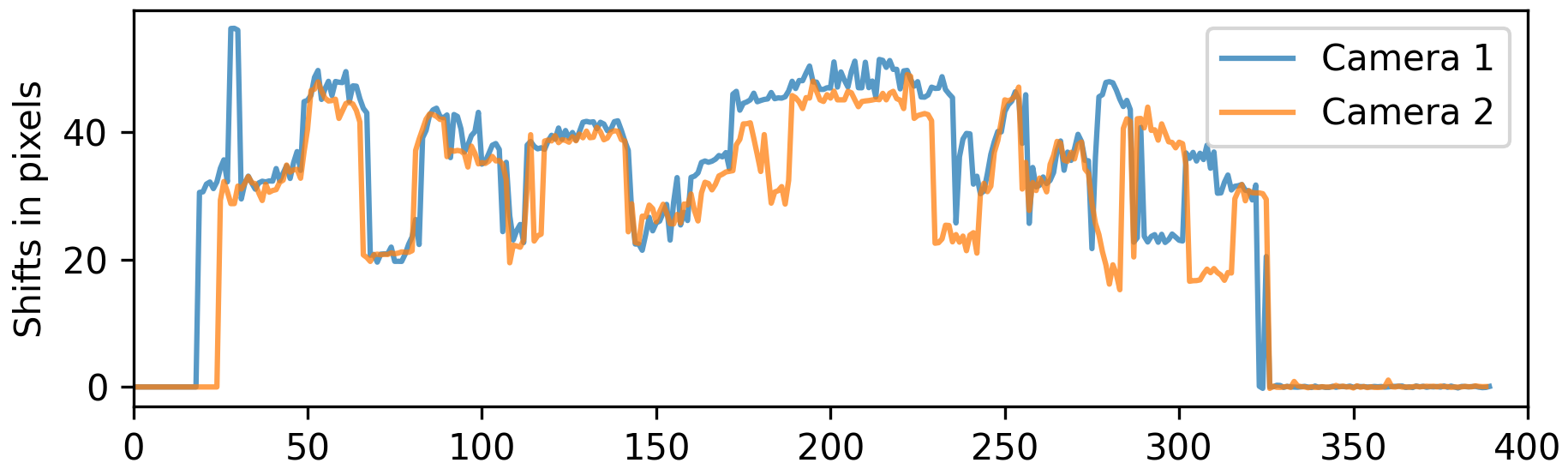}
         % \caption{The same shifts after syncing.}
         \caption{Shift alignment after synchronization.}
         \label{fig:sync2}
     \end{subfigure}
     \hfill
     \begin{subfigure}{\linewidth}
         \centering
         \includegraphics[width=\textwidth]{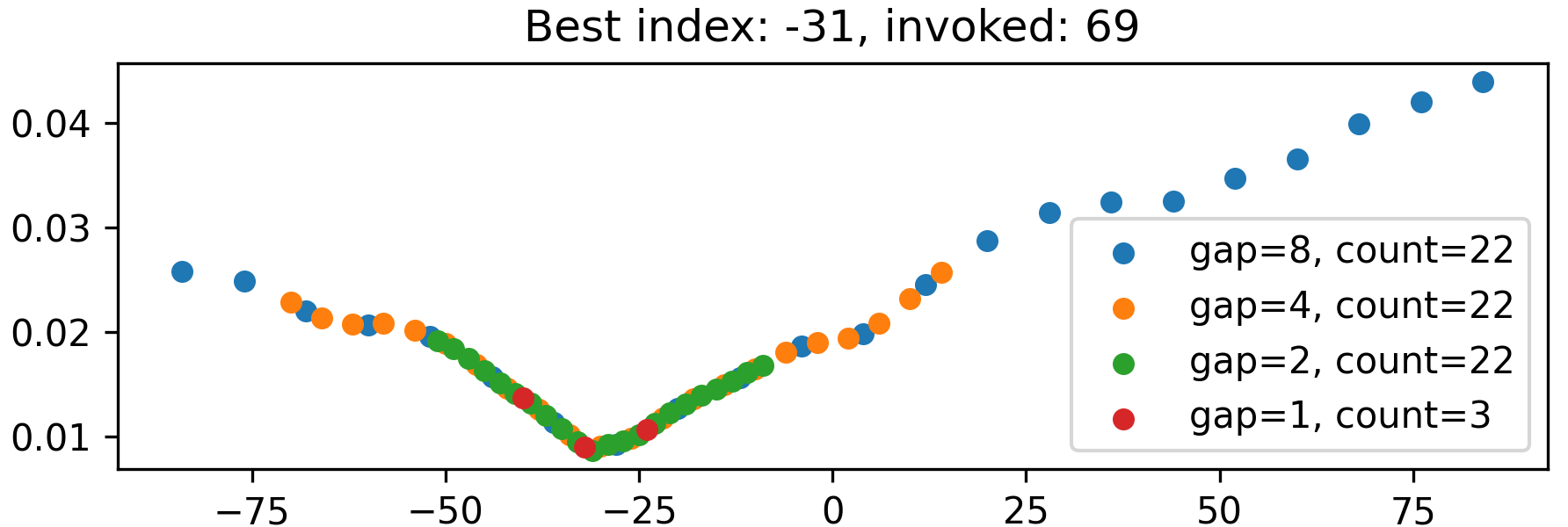}
         % \caption{Errors at various index offsets. Shows that it is mostly convex with minimum being the index that syncs the videos.}
         \caption{L1 loss at various offsets. The minimum point shows optimal synchronization.}
         \label{fig:sync3}
     \end{subfigure}
        % \caption{Video synchronization of two videos using phase correlation and optimizing L1 loss.}
        \caption{Video synchronization of two videos using phase correlation and L1 loss minimization.}
        \label{fig:sync}
    % \vspace{-0.25 cm}
\end{figure}

As we increase or decrease the offset, the error moves monotonically towards the minimum (Fig.~\ref{fig:sync3}), thus, we treat this as a convex optimization problem. To optimize the error, we follow an iterative method.
First, we explore uniformly within a fixed boundary of offsets shown in the blue dots in Fig.~\ref{fig:sync3}. Then, we pick the offset with the minimum error and explore within a smaller boundary shown in the orange dots. We repeat this process until we reach the minimum with no unexplored offsets within the boundary for that iteration.

For three videos, we use the center camera as the temporal reference and compute offsets for the left and right videos independently. Once aligned, we trim the excess frames to equalize video lengths, resulting in synchronized sequences of length $N_\text{synced}$.

\subsection{Structure-from-Motion (SfM)}
\label{sub:structure_from_motion}
We pass the synchronized videos and camera intrinsic parameters to our SfM pipeline to generate a point cloud that represents the vehicle's undercarriage and compute precise camera poses. This sparse point cloud will then serve as the seed for Gaussian splatting (Sec.~\ref{sub:gaussian_splatting}). Our SfM pipeline is specifically tailored to address the challenges of a low-parallax, wide-angle capture by integrating a robust, learning-based feature matching workflow with strong geometric priors from our camera rig. Key algorithmic choices are detailed below: 

\subsubsection{Frame Selection and Image Undistortion}
\label{subsub:frame_selection_and_masking}
Given the vehicle's speed and proximity to the cameras, motion blur can degrade image quality. To reduce computation and improve quality, we uniformly sample the sharpest $k = 250$ frame triplets from the synchronized left, center, and right camera videos. Sharpness is scored by the average Laplacian variance of the frame triplets (Eq.~\ref{eq:score_laplacian_variance}).

% \vspace{-0.25 cm}
\begin{equation}
\label{eq:score_laplacian_variance}
   \mathrm{Score}(I) =  \frac{1}{3} (\mathrm{Var}\bigl(\nabla^2 I_l\bigr) + \mathrm{Var}\bigl(\nabla^2 I_c\bigr) + \mathrm{Var}\bigl(\nabla^2 I_r\bigr) )
\end{equation}

\noindent
where $I_l$, $I_c$, and $I_r$ are the gray-scale images from the left, center, and right cameras, respectively, and $\nabla^2 I$ is the discrete Laplacian of the gray-scale image $I$.

A key step in our pipeline is the initial correction of severe lens distortion. We undistort each selected frame using the calibrated camera parameters obtained from our one-time ChArUco calibration process (Sec.~\ref{sub:camera_calibration}). This pre-processing step rectifies the images, ensuring that features are detected in a geometrically correct space. 

\subsubsection{Feature Extraction}
To enhance local contrast and reveal features in darker, underexposed areas of the undercarriage, we apply CLAHE, which adds $15 - 25\, \%$ more features. 

We utilize DISK, a learned local feature descriptor, for its robustness in challenging conditions. It is trained on a massive dataset of images to detect and describe features that are highly repeatable and reliable across significant viewpoint and illumination changes, which are common in our undercarriage capture environment. We extract up to $8192$ DISK features from each pre-processed frame.

\subsubsection{Constrained Feature Matching}
Standard SfM approaches often rely on exhaustive or sequential matching, which can be computationally expensive and prone to error in scenes with repetitive structures or challenging motion from multi-camera rigs. We leverage the known spatiotemporal structure of our capture process to implement a highly efficient and robust constrained matching strategy.

 For each image triplet at index $i$, we consider a temporal window $\mathcal{W}_{5}(i)=\{\,i-5,\dots,i+5\,\}$ and create three classes of pairs:
\begin{align*}
& \text{(i) intra-camera}  : (c,i) \leftrightarrow (c,j), \quad c \in \{\text{L,C,R}\}, & j \in \mathcal{W}_{5}(i), \\
& \text{(ii) cross L$\to$C}  : (\text{L},i) \leftrightarrow (\text{C},j), & j \in \mathcal{W}_{5}(i), \\
& \text{(iii) cross C$\to$R} : (\text{C},i) \leftrightarrow (\text{R},j), & j \in \mathcal{W}_{5}(i).
\end{align*}

We intentionally omit direct matching between the left~(L) and right~(R) camera views because the relatively large horizontal distance ($62\, \rm{cm}$ baseline) between these cameras, compared to the short vertical distance ($12-30\, \rm{cm}$) from the camera to the vehicle undercarriage, creates extreme perspective differences, making accurate feature matching difficult. The intra-camera matches help track camera motion over time, whereas the cross-camera matches provide the stereo baseline necessary for robust triangulation. The feature descriptors from these constrained pairs are matched using LightGlue, which uses an attention-based graph neural network to find correspondences. LightGlue considers the global context of all features in both images, allowing it to produce a highly accurate set of raw matches with few outliers. These raw matches are then imported into COLMAP, where a final geometric verification is performed using RANSAC \cite{schnabel2007efficient} to filter any remaining outliers and ensure consistency with our calibrated camera models. This two-stage approach, first constraining the search space spatiotemporally, then employing a high-fidelity learned matcher, synergistically accelerates the matching process while minimizing erroneous correspondences.

% Talk about guided matching and relative pose estimation.

\subsubsection{Rig-Aware Sparse Point Cloud Generation}
We use an incremental SfM method that starts with a strong image pair and progressively adds more views. A key part of our methodology is integrating the known camera rig geometry as a prior within the bundle adjustment (BA) optimization. 

After triangulating 3D points, BA refines the camera poses and 3D structure by minimizing the total reprojection error (Eq.~\ref{eq:bundle_adjustment}). The objective function is:

\begin{equation}
J = \min_{\mathbf{C}_i,\, \mathbf{P}_j} \sum_{i,j} w_{ij} \left\| \pi(\mathbf{C}_i, \mathbf{P}_j) - \mathbf{p}_{ij} \right\|^2
\label{eq:bundle_adjustment}
\end{equation}

\noindent where $\mathbf{C}_i$ is the pose of camera $i$, $\mathbf{P}_j$ is the 3D position of point $j$, $\mathbf{p}_{ij}$ is the observed projection of point $j$ in camera $i$, and $w_{ij}$ is a binary term indicating visibility. 

To enforce the known camera rig geometry, we augment this objective function with a regularization term based on the relative camera positions and orientations on our camera rig. As per our camera rig schematics, we provide a pose prior for the left and right cameras relative to the center camera.
% \vspace*{-0.5\baselineskip}
\[
\mathbf{t}_\text{L-C} = [-0.31, 0, 0]^\top, \quad \mathbf{t}_\text{C-R} = [+0.31, 0, 0]^\top
\]
% \vspace*{-1.75\baselineskip}

This rig-aware BA prevents inter-camera drift, resulting in an accurate sparse point cloud.

\subsection{Gaussian Splatting}
\label{sub:gaussian_splatting}
We pass the sparse point cloud generated from our SfM pipeline (Sec.~\ref{sub:structure_from_motion}) along with the undistorted images to the 3D Gaussian splatting framework introduced by Kerbl et al. \cite{kerbl20233d} to transform the discrete set of 3D points into a dense radiance representation that can be rendered from any viewpoint.

Each point in the sparse point cloud is initialized as a Gaussian:
% \vspace*{-0.5\baselineskip}
\begin{equation}
    G_i(\mathbf{x}) = \alpha_i \exp\Bigl[-\tfrac{1}{2}(\mathbf{x}-\boldsymbol{\mu}_i)^{\!\mathsf{T}} \mathbf{\Sigma}_i^{-1} (\mathbf{x}-\boldsymbol{\mu}_i) \Bigr]
\end{equation}
% \vspace*{-1.25\baselineskip}

\noindent where $\boldsymbol{\mu}_i \in \mathbb{R}^3$ is the mean position, $\alpha_i$ is its initial opacity, and $\boldsymbol{\Sigma}_i \in \mathbb{R}^{3\times3}$ is a diagonal covariance matrix scaled by the local point density.

The 3D Gaussian parameters are then optimized, yielding a dense radiance representation of the scene. For rendering, each 3D Gaussian $G_i$ is projected onto the image space as a 2D Gaussian $G_i(\mathbf{u}), \text{where,\,} \mathbf{u}=(u,v)$. Visibility and color are composited via front-to-back alpha blending:
% \vspace*{-0.75\baselineskip}
\begin{equation}
    \mathbf C(\mathbf u)
=\sum_{i=1}^N w_i(\mathbf u)\,\mathbf c_i,
\quad
w_i(\mathbf u)
=\tilde{\alpha}_i(\mathbf u)\prod_{j<i}\bigl[1-\tilde{\alpha}_j(\mathbf u)\bigr]
\end{equation}
% \vspace*{-0.75\baselineskip}

The SfM points are used to initialize the means $\boldsymbol{\mu}_i$, colors~$c_i$ are initialized based on the color of the associated image pixel, and covariances $\boldsymbol{\Sigma}_i$ are initialized as isotropic $\sigma^2\mathbf{I}$. 

% Training takes $\approx 8-10$ minutes on an RTX A6000 GPU. Fast training and real-time rendering enable vehicle condition inspectors and customers to view the vehicle's undercarriage from different angles, making it practical for production use.

\section{Experiments}
This section evaluates each stage of our pipeline and highlights the design choices that enable high-fidelity 3D Gaussian splatting. 

We tested our system using ten vehicles of varying makes, models, and undercarriage conditions (e.g., new, rusted, damaged), recorded with our wide-angle camera rig. Each vehicle's drive-through produces three videos (left, center, right) with a resolution of $1920 \times 1080$ at $120$ frames per second (fps). The videos are $\approx 8$ seconds long, giving us $\approx 960$ frames per camera.

All experiments were conducted on a workstation equipped with an AMD EPYC 7763 CPU, 2 TB of RAM, and eight NVIDIA RTX A6000 GPUs.

\subsection{Camera Calibration}
\label{sub:camera_calibration_results}
% Correcting the extreme distortion introduced by wide-angle lenses requires accurate camera calibration. An inaccurate camera model would propagate errors through the entire reconstruction pipeline, leading to warped point clouds and distorted renders. We evaluated three camera distortion models by fitting them to our curated set of sharp ChArUco board images.

% The eight-parameter Full OpenCV model, which accounts for higher-order radial distortion terms and tangential distortion, achieved the lowest RMS projection error: $0.74$ pixels. 

% \begin{table}[ht]
% \centering
% \caption{Comparison of camera calibration models. The Full OpenCV model yields the lowest RMS reprojection error.}
% \label{tab:calibration_models}
% % \begin{tabular}{|l|c|}
% \begin{tabular}{lc}
% \hline
% \textbf{Distortion Model} & \textbf{RMS Error (pixels)} \\
% \hline
% 4-Parameter Fisheye        & 1.26 \\
% \hline
% 5-Parameter Standard       & 0.93 \\
% \hline
% \textbf{8-Parameter Full OpenCV} & \textbf{0.74} \\
% \hline
% \end{tabular}
% % \vspace{-0.25 cm}
% \end{table}

Correcting the extreme distortion introduced by wide-angle lenses requires accurate camera calibration. An inaccurate camera model would propagate errors through the entire reconstruction pipeline, leading to warped point clouds and distorted renders. We evaluated the Full OpenCV camera model by fitting it to our curated set of sharp ChArUco board images. The eight-parameter Full OpenCV model, which accounts for higher-order radial distortion terms and tangential distortion, achieved a low RMS projection error of $0.74$ pixels.

\begin{figure}[t]
    \centering
    \includegraphics[width=0.99\linewidth]{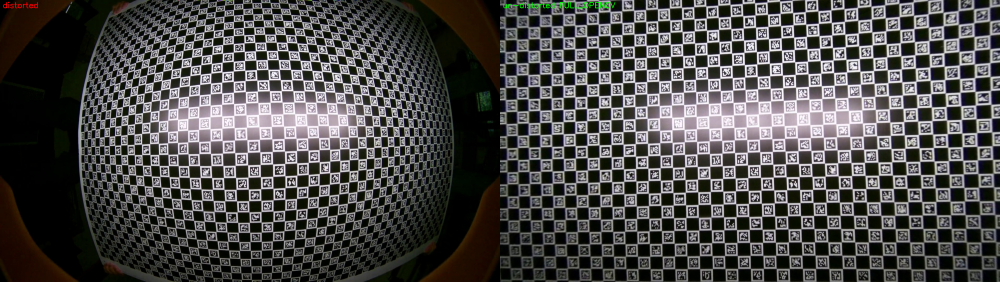}
    \caption{Qualitative comparison of original calibration image (left) and undistorted image using Full OpenCV model (right).}
    \label{fig:charuco_calibration_comparison}
    % \vspace{-0.25 cm}
\end{figure}

\begin{figure}[t]
    \centering
    \includegraphics[width=0.99\linewidth]{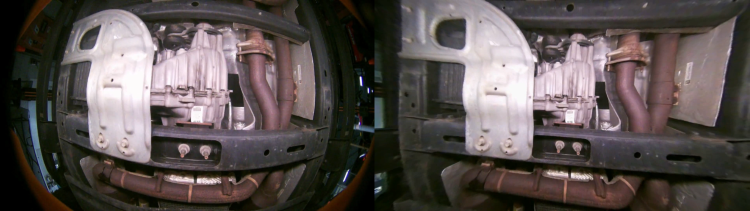}
    \caption{Qualitative comparison of original undercarriage image (left) and undistorted image using Full OpenCV model (right).}
    \label{fig:undistortion_example_undercarriage}
    % \vspace{-0.25 cm}
\end{figure}

Fig.~\ref{fig:charuco_calibration_comparison} shows how the Full OpenCV camera model corrects severe distortion in calibration images. Fig.~\ref{fig:undistortion_example_undercarriage} shows the same correction on a real undercarriage image. Accurate intrinsic parameters minimize bundle-adjustment drift in the SfM pipeline, resulting in geometrically accurate point clouds.

\subsection{Video Synchronization}
Even with a hardware trigger, minor latencies can lead to temporal misalignment between the three video streams. Our two-stage synchronization algorithm is designed to correct these offsets with sub-frame precision. 

The videos from the three cameras could be out of sync by up to $\pm\, 35$ frames. After synchronization, the average difference in vertical motion between camera pairs drops from $774$ pixels to $\approx 22$ pixels.

The algorithm uses a convex loss function, which helps reliably find the best alignment. Precise synchronization ensures spatial-temporal correspondence for triplet selection, thereby preventing ghost artifacts in the Gaussian splat.

\subsection{Structure-from-Motion}
\label{sub:structure_from_motion_results}
The SfM stage generates the geometric foundation for our model. It outputs a sparse 3D point cloud and camera poses, which set an upper bound on the quality of the Gaussian splat. To validate our methodology, we perform a comprehensive evaluation comparing our proposed pipeline against two key baselines: a vanilla SfM approach and a version of our rig-aware pipeline using classic SIFT features and COLMAP's matcher instead of DISK+LightGlue. We then conduct a series of ablation studies on our pipeline to demonstrate the impact of each of our core contributions.

The quantitative results are detailed in Table~\ref{tab:sfm_ablation}. The vanilla Baseline SfM, which does not include our calibration, synchronization, custom matching strategy, and rig-based pose priors, fails to produce a coherent reconstruction. Our rig-aware pipeline using classic SIFT features generates a strong baseline, significantly improving all metrics. By integrating learned features (DISK) and an attention-based matcher (LightGlue), our proposed methodology achieves the best overall performance, yielding the densest and most accurate sparse point cloud. The ablation studies highlight the importance of each component in our final pipeline. Removing either the dedicated camera calibration (and thus pre-undistortion), the custom matching strategy, or the rig-based pose priors results in degenerate point clouds that do not represent the vehicle undercarriage, and thereby result in degenerate Gaussian splats as shown in Fig.~\ref{fig:degenerate_gaussian_splats}.

\begin{table}[ht]
% \vspace{-0.15 cm}
\centering
\caption{SfM baseline comparison and ablation study. Each ablation removes a specific component from our SfM pipeline.}
\label{tab:sfm_ablation}
\setlength{\tabcolsep}{3 pt}
\renewcommand{\arraystretch}{1.2}
  \resizebox{\columnwidth}{!}{
% \begin{tabular}{|l|c|c|c|c|c|c|c|}
\begin{tabular}{lccccccc}
\hline
\textbf{Metric} & 
\begin{tabular}[c]{@{}c@{}}\textbf{Baseline} \\ \textbf{SfM*}\end{tabular} &
\begin{tabular}[c]{@{}c@{}}\textbf{Rig-Aware} \\ \textbf{SfM (SIFT)}\end{tabular} & 
\begin{tabular}[c]{@{}c@{}}\textbf{- Camera} \\ \textbf{Calibration*}\end{tabular} & 
\begin{tabular}[c]{@{}c@{}}\textbf{- Video} \\ \textbf{Sync}\end{tabular} & 
\begin{tabular}[c]{@{}c@{}}\textbf{- Custom} \\ \textbf{Matching*}\end{tabular} & 
\begin{tabular}[c]{@{}c@{}}\textbf{- Pose} \\ \textbf{Priors*}\end{tabular} & 
\begin{tabular}[c]{@{}c@{}}\textbf{Our SfM} \\ \textbf{DISK+LG}\end{tabular} \\
\hline
\begin{tabular}[c]{@{}l@{}}Registered \\ Images ↑\end{tabular}  & 749 & \textbf{750} & \textbf{750} & \textbf{750} & 745 & 673 & \textbf{750} \\
\hline
\begin{tabular}[c]{@{}l@{}}Sparse 3D \\ Points ↑\end{tabular}  & 126,496 & 323,005 & 189,121 & 339,209 & 368,751 & 313,395 & \textbf{427,299} \\
\hline
\begin{tabular}[c]{@{}l@{}}Mean Track \\ Length ↑\end{tabular} & \textbf{10.0868} & 8.6509 & 6.8651 & 8.1906 & \textbf{14.0779} & 8.0244 & \textbf{8.6674} \\
\hline
\begin{tabular}[c]{@{}l@{}}Reprojection \\ Error (px) ↓\end{tabular} & 0.7817 & 0.5376 & 0.7058 & 0.4995 & 0.5279 & 0.5478 & \textbf{0.4909} \\
\hline
\end{tabular}}
\caption*{\footnotesize * These SfM approaches yield a degenerate sparse point cloud, which doesn't represent a vehicle undercarriage. Consequently, these sparse point clouds produce a degenerate Gaussian splat.}
% \vspace{-0.35 cm}
\end{table}

\begin{figure}[t]
    \centering
    \includegraphics[width=0.99\linewidth]{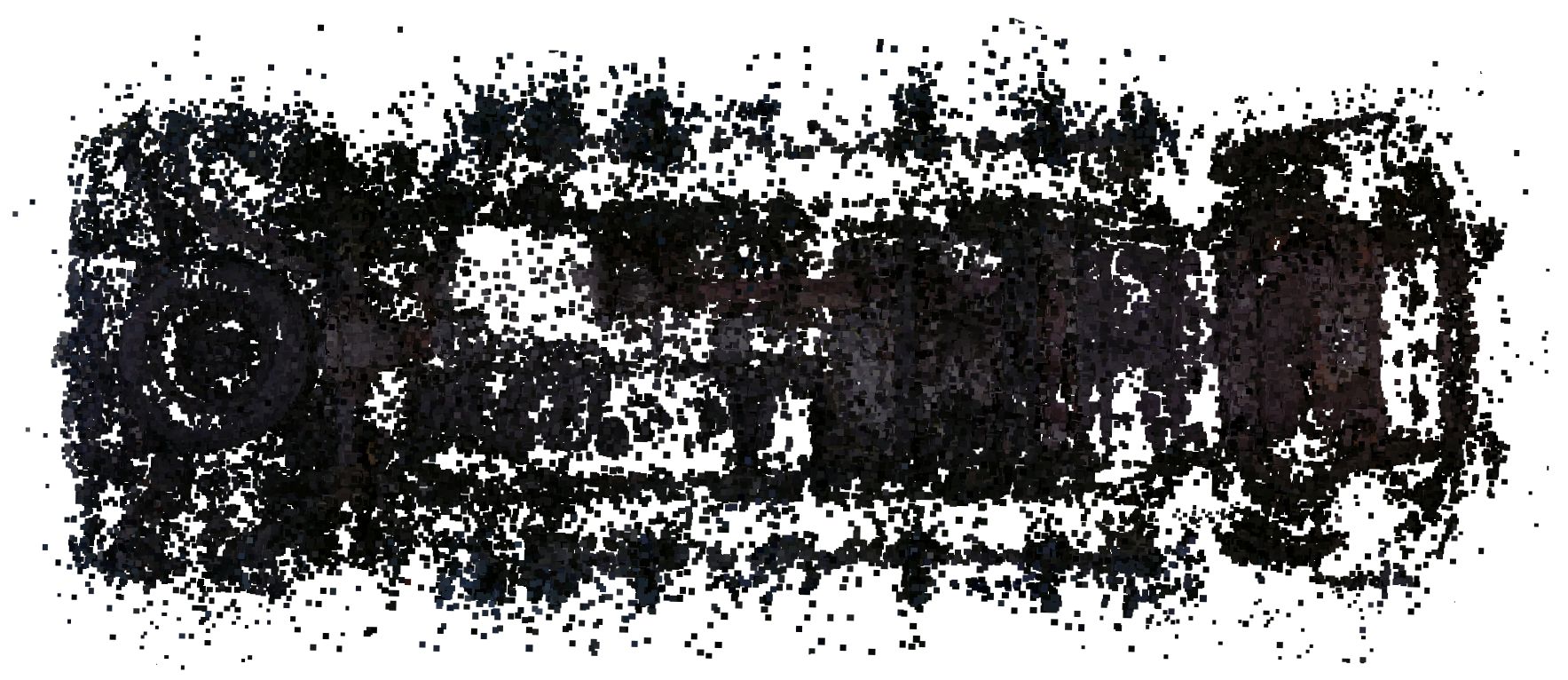}
    \caption{Sparse point cloud for a vehicle's undercarriage.}
    \label{fig:sparse_point_cloud}
    % \vspace{-0.35 cm}
\end{figure}

As shown in Fig.~\ref{fig:sparse_point_cloud}, our final method generates a dense, coherent representation of the undercarriage, providing a solid geometric base for Gaussian splatting.

\subsection{Gaussian Splatting}
\label{sub:gaussian_splatting_results}

Finally, we evaluate the Gaussian splatting results of the 3D vehicle undercarriage. To evaluate our results, we use standard image quality metrics: Peak Signal-to-Noise Ratio (PSNR), Structural Similarity Index (SSIM), and the perceptual Learned Perceptual Image Patch Similarity (LPIPS) metric.

\begin{figure}[t]
    \centering
    \includegraphics[width=0.95\linewidth]{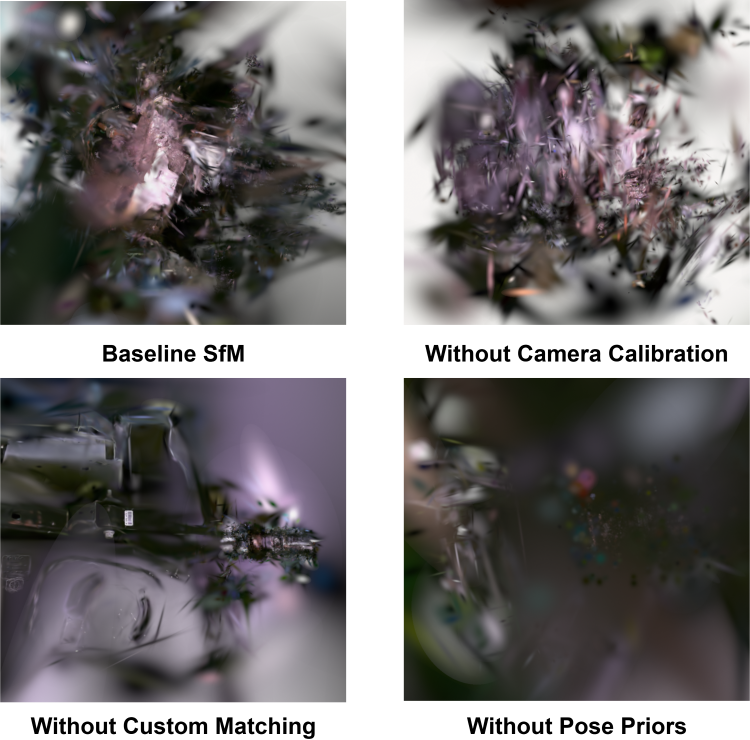}
    \caption{The baseline SfM and ablations for camera calibration, custom matching, and pose priors all result in degenerate, noisy, or incomplete renders. These results visually confirm the quantitative findings in Table \ref{tab:gs_ablation} and demonstrate that each component of our proposed pipeline is essential for generating a coherent and high-fidelity 3D model.}
    \label{fig:degenerate_gaussian_splats}
    % \vspace{-0.5 cm}
\end{figure}

As detailed in Table \ref{tab:gs_ablation}, our pipeline produces good results, with a mean PSNR of $30.66$ dB, a mean SSIM of $0.92$, and a mean LPIPS of $0.19$, indicating that our rendered views are visually and perceptually closer to the real camera images. This demonstrates a clear improvement over a version of our pipeline that uses classic SIFT features and matcher, indicating that the denser and more accurate point cloud from the DISK+LightGlue approach provides a superior seed for the Gaussian splatting optimization. As visualized in Fig.~\ref{fig:degenerate_gaussian_splats}, the baseline and several ablated SfM initializations result in degenerate, noisy, or incomplete renders, confirming that our pipeline is crucial for achieving a high-quality 3D model.

\begin{figure}[t]
    \centering
    \includegraphics[width=0.99\linewidth]{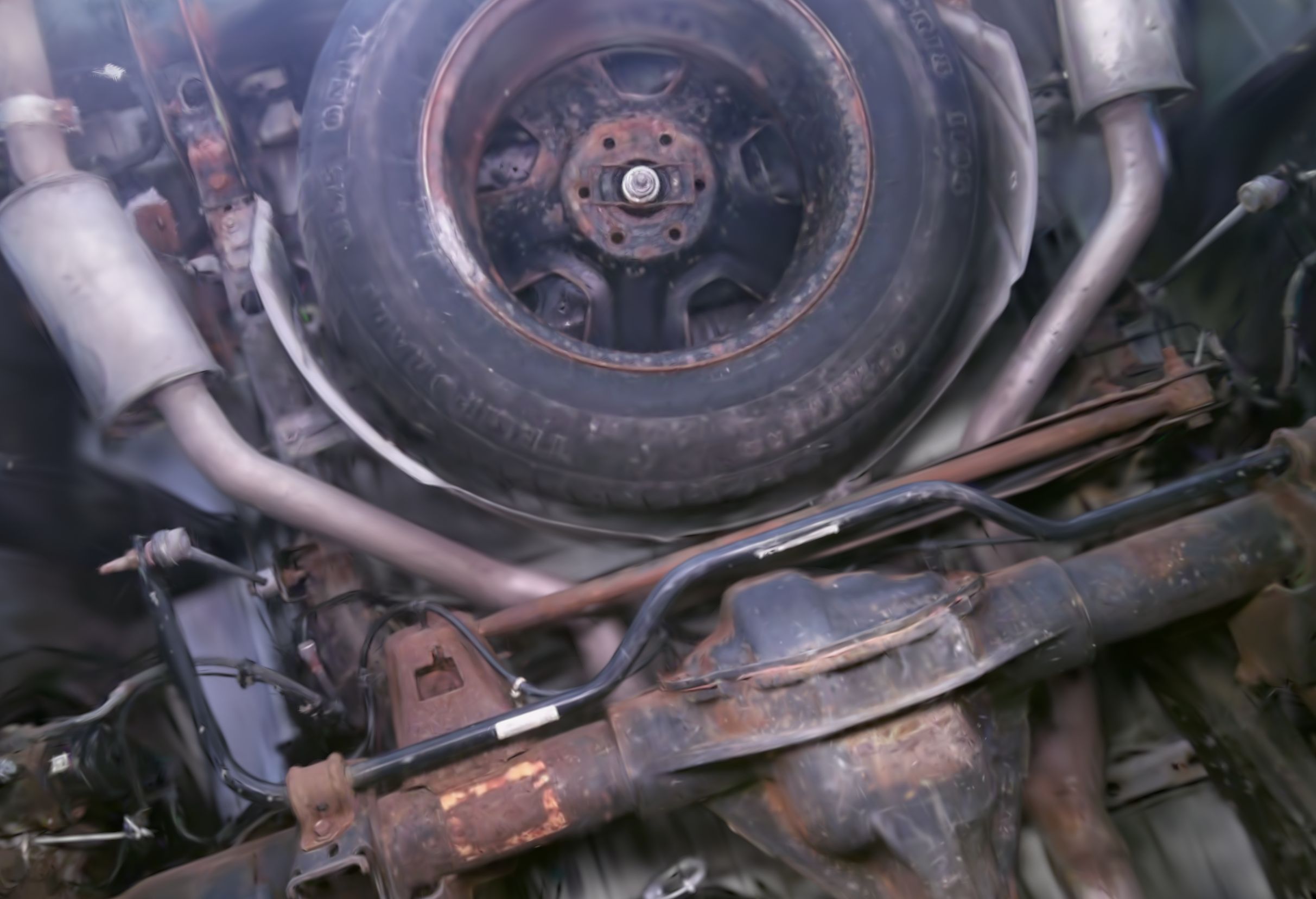}
    \caption{Gaussian splat render highlighting the ability of our pipeline to capture fine-grained textures and diagnostic details. The render shows areas of significant rust on the frame. This level of detail enables accurate vehicle condition assessments by inspectors and increases buyer confidence in online marketplaces.}
    \label{fig:gaussian_splat_rusty_undercarriage}
    % \vspace{-0.25 cm}
\end{figure}

\begin{figure}[t]
    \centering
    \includegraphics[width=0.99\linewidth]{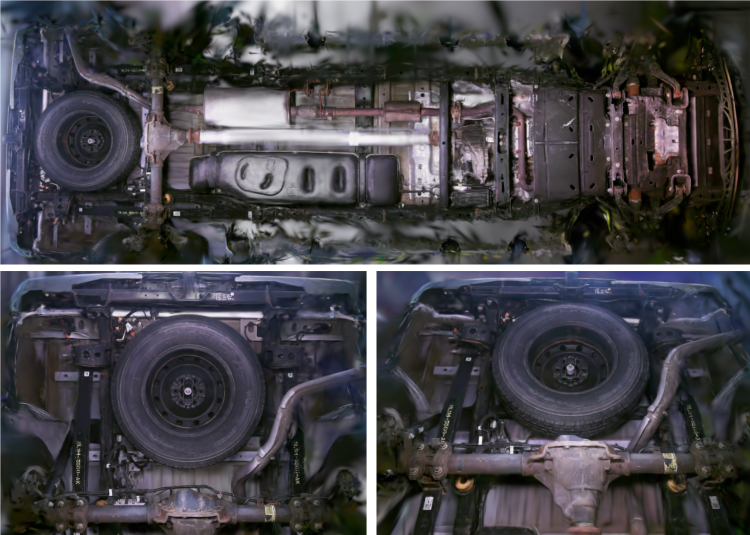}
    \caption{Photorealistic 3D undercarriage models produced by our pipeline. The top image shows a comprehensive overhead view of the vehicle's undercarriage, while the bottom images provide different perspectives of the rear.}
    \label{fig:gaussian_splat_results}
    % \vspace{-0.35 cm}
\end{figure}

Our method achieves real-time rendering performance, averaging over $130$ frames per second (fps), making it suitable for interactive inspection. Fig.~\ref{fig:gaussian_splat_results} shows that our models are photorealistic and geometrically coherent. The high-quality SfM initialization allows the 3D-GS optimization to converge to a sharp representation, capturing fine details such as bolt heads, rust patterns, fluid marks, and the metallic sheen of exhaust systems (Fig. \ref{fig:gaussian_splat_rusty_undercarriage}). Notably, the models are largely free of the ``floater" artifacts and hazy regions that often plague radiance fields trained from less accurate camera poses. 

Training takes $\approx 8-10$ minutes on an RTX A6000 GPU. Efficient training and a high-fidelity 3D model enable us to view the vehicle's undercarriage from different angles, identifying potential issues such as corrosion, leaks, or damage, making it practical for production use.

\begin{table}[ht]
\centering
\caption{Gaussian splatting baseline comparison and ablation study. Each ablation removes a specific component from our SfM pipeline.}
\label{tab:gs_ablation}
\setlength{\tabcolsep}{3pt}
\renewcommand{\arraystretch}{1.2}
  \resizebox{\columnwidth}{!}{
% \begin{tabular}{|l|c|c|c|c|c|c|c|}
\begin{tabular}{lccccccc}
\hline
\textbf{Metric} & 
\begin{tabular}[c]{@{}c@{}}\textbf{Baseline} \\ \textbf{SfM*}\end{tabular} &
\begin{tabular}[c]{@{}c@{}}\textbf{Rig-Aware} \\ \textbf{SfM (SIFT)}\end{tabular} & 
\begin{tabular}[c]{@{}c@{}}\textbf{- Camera} \\ \textbf{Calibration*}\end{tabular} & 
\begin{tabular}[c]{@{}c@{}}\textbf{- Video} \\ \textbf{Sync}\end{tabular} & 
\begin{tabular}[c]{@{}c@{}}\textbf{- Custom} \\ \textbf{Matching*}\end{tabular} & 
\begin{tabular}[c]{@{}c@{}}\textbf{- Pose} \\ \textbf{Priors*}\end{tabular} & 
\begin{tabular}[c]{@{}c@{}}\textbf{Our SfM} \\ \textbf{DISK+LG}\end{tabular} \\
\hline
PSNR (dB) ↑  & 23.46 & 29.68 & 21.40 & 29.47 & 20.78 & 16.96 & \textbf{30.66} \\
\hline
SSIM ↑       & 0.78 & 0.91 & 0.75 & 0.89 & 0.76 & 0.68 & \textbf{0.92} \\
\hline
LPIPS ↓      & 0.42 & 0.21 & 0.57 & 0.26 & 0.43 & 0.66 & \textbf{0.19} \\
\hline
\end{tabular}}
% \vspace{-0.35 cm}
\end{table}

\section{Conclusion}
In this paper, we presented a complete pipeline for generating high-quality, interactive 3D models of vehicle undercarriages using multi-view video. Our system is designed to address the practical challenges of under-vehicle inspection, which is often tedious, physically demanding, and especially difficult to scale for online marketplaces. By converting raw videos into detailed, photorealistic 3D models, our method enables faster, safer, and more transparent vehicle assessments.

The core of our approach is a rig-aware SfM pipeline, built specifically to support 3D Gaussian splatting. We found that accurately handling wide-angle lens distortion and utilizing physical priors from the camera rig are key to recovering reliable geometry in challenging settings, such as low parallax and linear motion. This solid geometric foundation enabled the generation of sharp, artifact-free radiance fields.

Our experiments show that the pipeline performs well, producing high-quality renderings (PSNR $> 30$, SSIM $> 0.90$) at real-time speeds ($>130$ FPS). The resulting 3D models reveal diagnostic features like rust, leaks, and surface wear, making the tool valuable for vehicle inspectors and consumers. We see this work as a foundation for future research in automated 3D damage detection, with potential applications in other industrial inspection settings.

% \section*{Acknowledgment}
% This material is based upon work supported by the National Science Foundation under Grant Nos. CNS-2413876 and CNS-2120369. The authors thank ACV Auctions for providing data, hardware, and computing resources.

\bibliographystyle{IEEEtran}
\bibliography{references}

\end{document}